\documentclass[lettersize,journal]{IEEEtran}
\usepackage{amsmath,amsfonts}
\usepackage{algorithmic}
\usepackage{algorithm}
\usepackage{array}
\usepackage[caption=false,font=footnotesize,labelfont=sf,textfont=sf]{subfig}
\usepackage{textcomp}
\usepackage{stfloats}
\usepackage{url}
\usepackage{verbatim}
\usepackage{graphicx}
\usepackage{cite}
\usepackage{pifont}
\newcommand{\sfig}[1]{\raisebox{0.4pt}{\textcircled{\raisebox{-0.7pt}{\scriptsize #1}}}}

\newcommand{\cmark}{\ding{51}}
\newcommand{\xmark}{\ding{55}}

\begin{document}

\title{GHySSeg: Real-Time Geometric Ground and Road Segmentation for Hybrid Solid-State LiDAR}

\author{Martín Palos,~\IEEEmembership{Student Member,~IEEE,} and Fernando García,~\IEEEmembership{Member,~IEEE}
\thanks{The authors are with the Autonomous Mobility and Perception Lab (AMPL), Department of Systems Engineering and Automation, Universidad Carlos III de Madrid, 28911 Legan\'es, Madrid, Spain (e-mail: 100345948@alumnos.uc3m.es; fegarcia@ing.uc3m.es).}%
}

\markboth{Palos and García: GHySSeg: Real-Time Geometric Ground and Road Segmentation for Hybrid Solid-State LiDAR}{Palos and García: GHySSeg: Real-Time Geometric Ground and Road Segmentation for Hybrid Solid-State LiDAR}

\IEEEpubid{\begin{minipage}{\textwidth}\centering\footnotesize\textbf{This work has been submitted to the IEEE for possible publication. Copyright may be transferred without notice, after which this version may no longer be accessible.}\end{minipage}}

\maketitle

\begin{abstract}
Hybrid solid-state LiDAR sensors are making their way into production vehicles, yet the data they produce differs structurally from that of the established spinning mechanical sensors behind today's autonomous-driving datasets: it is ring-free, dense, cluttered, and concentrated in a frontal field of view. The classical geometric methods that autonomous vehicles rely on for drivable space computation and navigation assume the ordered ring-and-column structure of a 360-degree spinning LiDAR and degrade or fail on this data, while learned alternatives are computationally heavy, hard to certify, and trained almost exclusively on spinning-LiDAR data. We demonstrate that the same dense, per-curve geometry is instead an opportunity to close the divide between the industry's preferred sensors and autonomous-driving research. We present GHySSeg, a purely geometric method that segments ground and drivable road from a single hybrid solid-state LiDAR frame by tracing each scan curve and reading two geometric signatures, a step in height and a fold-back in range, with no training and no graphics processing unit. To evaluate it, we release HySoLiD-Drive, the first public benchmark dedicated to ground and road segmentation on an automotive hybrid solid-state LiDAR, comprising 100 manually labeled and 7000 unlabeled frames recorded by two vehicles in real driving. On this benchmark, GHySSeg surpasses both geometric and learned state-of-the-art baselines across ground and road segmentation, reaching an intersection-over-union of 0.907 for road and 0.896 for ground at over 0.96 precision, while running in real time at about 22 milliseconds per frame on a single core.
\end{abstract}


\section{Introduction}
\label{sec:introduction}
\IEEEPARstart{L}{iDAR} (light detection and ranging) is the cornerstone of modern autonomous driving (AD). As an active sensor, it delivers rich three-dimensional (3-D) measurements that stay reliable under poor illumination, even in complete darkness, and that are far denser than those of radar sensors, enabling object detection and classification directly in 3-D for a thorough understanding of the space around the vehicle. Its long range, moreover, lets a vehicle anticipate the road ahead, even at high speed~\cite{lidar_ad_survey}.

\begin{figure}[t]
  \centering
  \includegraphics[width=\columnwidth]{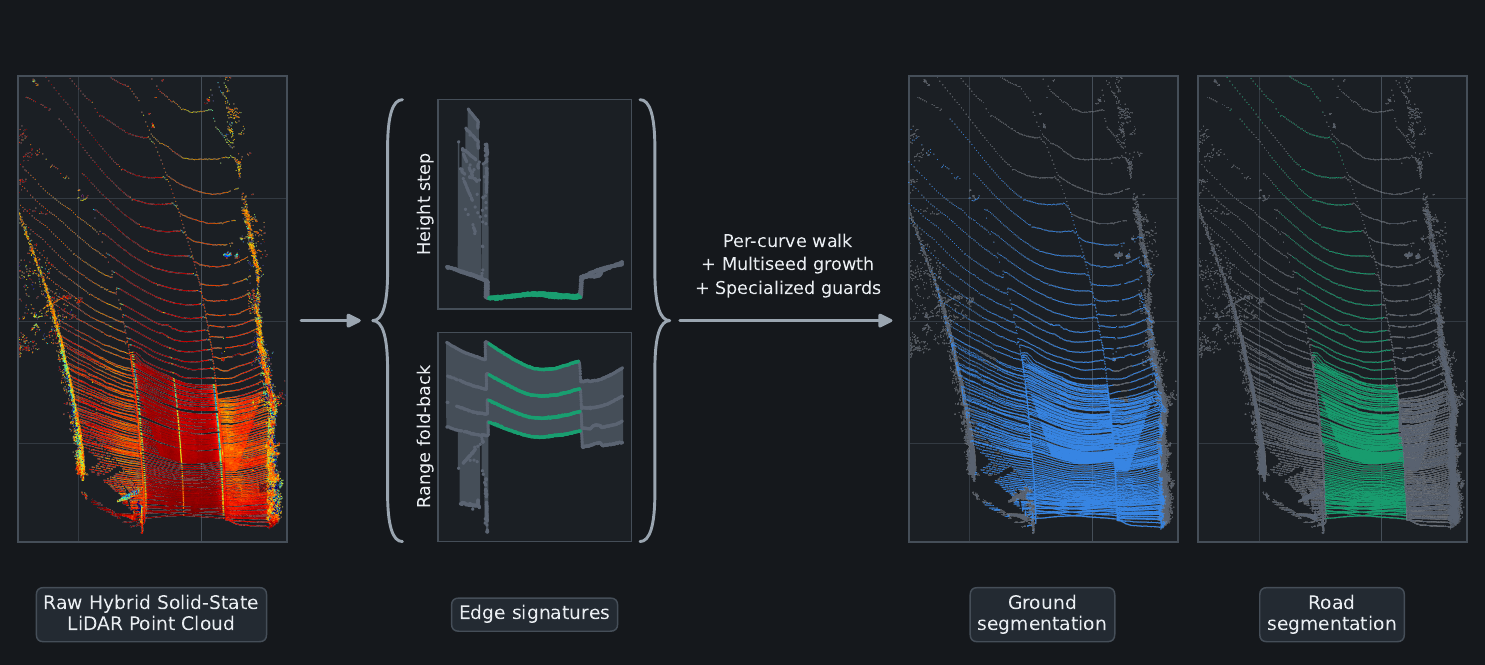}
  \caption{Overview of GHySSeg. From a single raw hybrid solid-state LiDAR frame, the method reads two geometric road-edge signatures (a step in height and a fold-back in range) and, through a per-curve walk with multiseed growth and specialized guards, segments the ground and the drivable road in real time.}
  \label{fig:overview}
\end{figure}

Spinning mechanical LiDARs, which provide a full $360^\circ$ field of view (FOV), dominate research: they are a mature, well-understood technology, and the public AD datasets and benchmarks are built almost exclusively on them~\cite{lidar_ad_survey}. Their adoption in consumer vehicles, however, faces real barriers. They are expensive; exploiting their $360^\circ$ coverage requires mounting them protruding above the roof, which is neither aerodynamic nor aesthetic and noticeably increases the vehicle's height clearance; and they rely on moving parts spinning at high speed.

\IEEEpubidadjcol These limitations have already driven a shift in the automotive industry toward solid-state (SS) and hybrid solid-state (HSS) LiDARs. Such sensors sacrifice full $360^\circ$ coverage in exchange for fewer (or no) moving parts, a higher and more concentrated point density, a cleaner and more seamless installation, and, above all, a lower cost~\cite{solid_state}. Recent consumer vehicle models, led predominantly by Chinese manufacturers such as Xiaomi, BYD, and NIO, with Western manufacturers such as Mercedes-Benz now following, already integrate SS or HSS LiDARs~\cite{xiaomi_su7, consumer_lidar}, bringing LiDAR to mass-production vehicles at last, beyond research platforms and robotaxis.

\begin{figure}[t]
  \centering
  \subfloat[Velodyne VLP-32C (mechanical)]{%
    \includegraphics[angle=90,width=0.47\columnwidth]{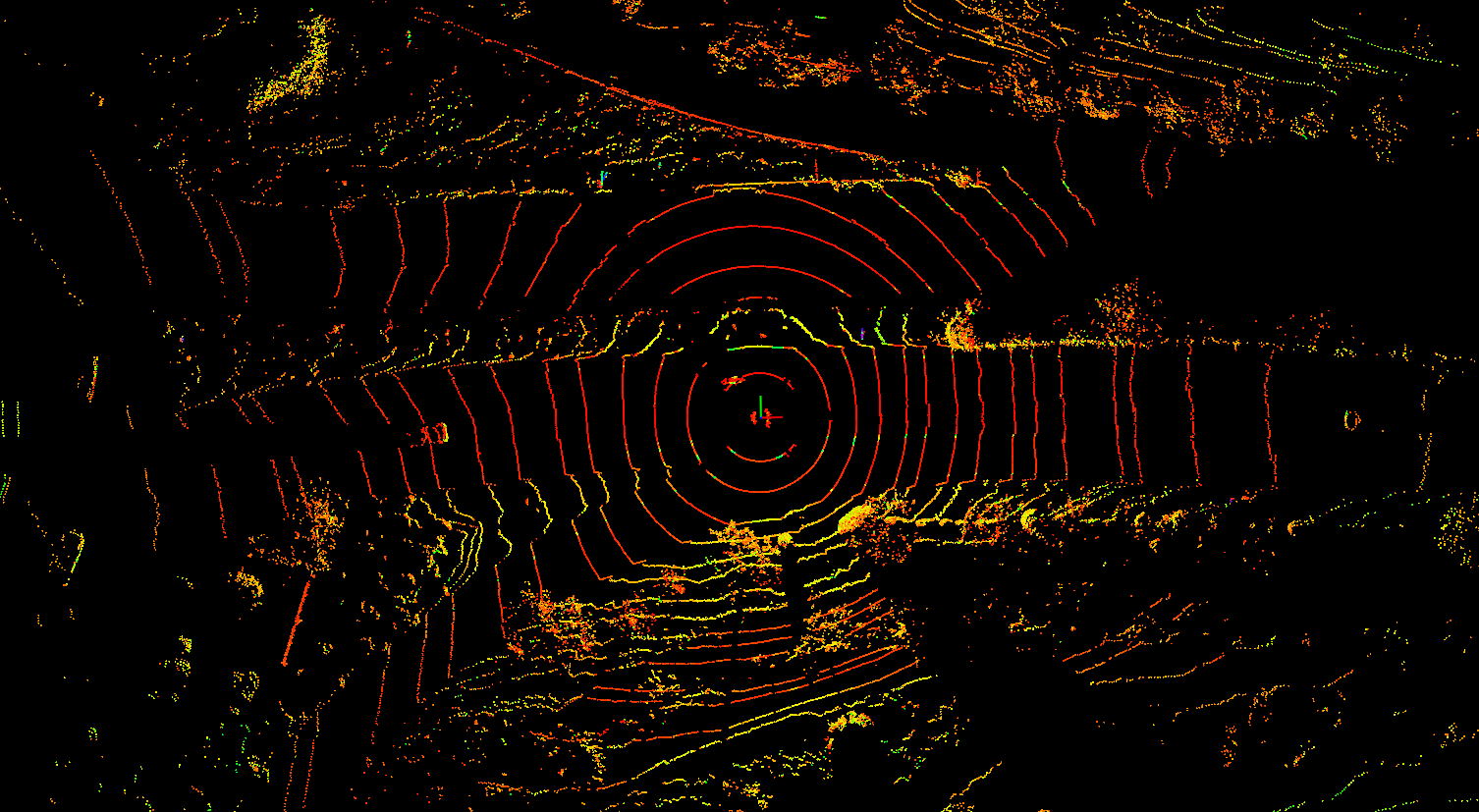}%
    \label{fig:pc-velodyne}}
  \hfil
  \subfloat[Seyond Falcon Kinetic (HSS)]{%
    \includegraphics[angle=90,width=0.47\columnwidth]{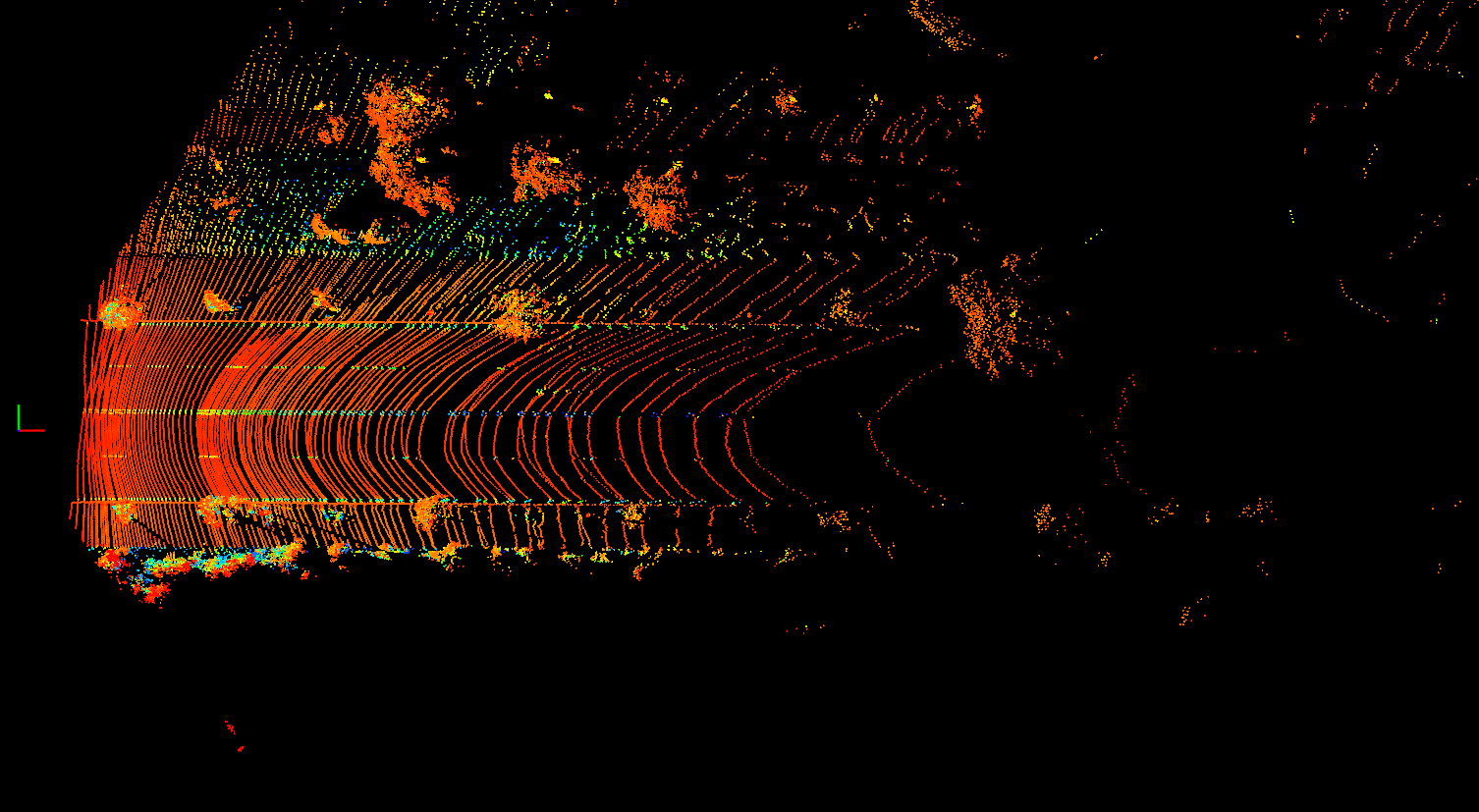}%
    \label{fig:pc-falcon}}
    \caption{Top-view comparison of point clouds from a mechanical LiDAR (a) and a hybrid solid-state LiDAR (b) in a real-world driving scenario.}
  \label{fig:pc-compare}
\end{figure}

This new technology, however, brings a challenge of its own: the data it produces is structurally different from that of a spinning sensor, as depicted in Fig.~\ref{fig:pc-compare}. HSS point clouds lack the structured ring pattern and the neatly ordered azimuth columns that spinning LiDARs owe to their rotating laser array, and, because the field of view is condensed into a frontal region, they are far denser and considerably less organized.

When facing this seemingly disorganized data, the classical geometric methods that autonomous vehicles rely on for perception, scene-understanding, and path-planning, such as ground filtering, drivable-area and road segmentation, and free-space estimation, degrade or fail outright, because they depend on the ordered ring-and-column structure of a spinning sensor. With the growing diversity of LiDAR manufacturers and models, the field's dominant answer is learned semantic segmentation~\cite{lidar_semseg_review,seg_survey}, in which a neural network labels every point directly from the raw cloud. These models, however, require a graphics processing unit (GPU) for inference that adds cost and energy demand, they behave unpredictably in unseen scenarios, offer little traceability of their errors, and, most importantly, they are trained almost exclusively on spinning-LiDAR datasets such as KITTI~\cite{kitti, semantickitti}. This introduces a domain gap: a model trained on cleanly spaced $360^\circ$ data struggles to generalize to the denser, frontal, more cluttered clouds of a solid-state sensor, even though both are 3-D point clouds~\cite{triess2021survey}.

Yet the very properties that break these methods are also an opportunity. The high density, the focused FOV, and the close, non-ring scan lines let an HSS sensor resolve surfaces, curbs, and centimetric steps sharply. In this dense per-curve structure, road boundaries leave a clean geometric signature: a step in height together with a visible bend in the range profile.

This article turns that observation into a method. We propose GHySSeg (Fig.~\ref{fig:overview}), which exploits the specific characteristics of automotive HSS LiDAR data to segment ground and road by pure geometry, without training or running heavy models, without a GPU, and in real time on a consumer-grade CPU, providing reliable and fast drivable-area 3-D detection for demanding autonomous-driving applications.

Because no existing public benchmark specifically targets ground and road segmentation on an automotive hybrid solid-state LiDAR as the main sensor, GHySSeg has been developed and validated on an Innovusion (now Seyond) Falcon Kinetic LiDAR~\cite{falcon_kinetic} mounted on two distinct autonomous vehicles operated by our laboratory at the Universidad Carlos III de Madrid (UC3M). Moreover, to directly address this automotive hybrid solid-state LiDAR data scarcity, we release HySoLiD-Drive~\cite{hysolid_drive_dataset} as part of this work. HySoLiD-Drive is an open benchmark of 100 unique, manually per-point-labeled frames, spanning urban roads, motorways, roundabouts, dense traffic, and sharp turns, together with 7000 unlabeled frames drawn from two continuous drives through Madrid, Spain. It comprises the data we have used to evaluate GHySSeg, and invites the community to build on it and benchmark against our results through an open, transparent leaderboard.

The contributions of this work are the following.
\begin{itemize}
    \item A sensor-native, real-time, single-shot, geometry-only ground and road segmentation method for HSS LiDAR.
    \item A per-curve road-boundary formulation that exploits the density and geometry of HSS data through height-step and range-fold-back signatures.
    \item HySoLiD-Drive, the first public benchmark dedicated to ground and road segmentation on an automotive hybrid solid-state LiDAR in real-world driving scenes, released with an open leaderboard.
    \item A real-world evaluation against six baselines with cross-vehicle transfer, displaying state-of-the-art accuracy at real-time CPU cost.
\end{itemize}

\section{Related Work}
\label{sec:related}

Ground and road segmentation are long-standing prerequisites for LiDAR perception within the autonomous driving field. In classical modular pipelines, they form the first stage, feeding everything in the AD stack from free-space estimation and real-time mapping to occupancy-grid construction and reactive obstacle detection~\cite{gomes2023survey}. Two schools of thought address them: geometric, heuristic methods and learned, machine-learning (ML) approaches. Their balance is task-dependent: on ground segmentation the two remain closely matched on public benchmarks, whereas on road segmentation the learned school clearly dominates. We therefore review and benchmark both schools on each task.

\subsection{Geometric Ground Segmentation}

Ground segmentation is the first stage of most non-ML LiDAR-centered perception pipelines: quickly discarding the measurements that belong to the ground is what makes the subsequent clustering and object detection possible and robust. Its classical solutions are purely geometric. Early methods fit a single ground plane by random sample consensus (RANSAC) or least squares, or discretize the scan into a polar or Cartesian elevation grid and threshold the local height~\cite{himmelsbach}. To handle non-planar terrain, approaches such as that of Zermas \emph{et al.} divide the cloud into radial bins and, for each bin, estimate the shape of the ground by fitting several connected straight-line segments extending away from the sensor~\cite{gpf}. Widely adopted open autonomous-driving stacks follow the same principle; Autoware, for instance, ships a ray-ground filter that walks each radial ray outward from the vehicle and labels points as ground from their height and angle~\cite{autoware}. More recent work improves robustness on sloped and uneven ground: Patchwork~\cite{patchwork} and its successor Patchwork++~\cite{patchworkpp} divide the scan into concentric zones, tailored to the ring structure of a spinning LiDAR, and fit a region-wise plane in each. Patchwork++ is one of the most widely adopted real-time geometric methods, reporting a ground-segmentation F1 score above 96\% on SemanticKITTI. A different geometric strategy is the cloth simulation filter (CSF)~\cite{csf}, which drapes a virtual cloth over the inverted cloud to separate ground from non-ground; originally developed for airborne and mapping LiDAR, it is robust but comparatively slow. Some recent methods even hybridize the two schools, pairing a geometric polar partitioning with a sparse convolutional network that resolves only the ambiguous, data-sparse regions~\cite{hybrid_ground}. These geometric methods were designed for spinning sensors: their radial bins, concentric zones, and elevation grids all assume the ordered ring-and-azimuth sampling of a rotating LiDAR. Moreover, their limited precision at small elevations such as curbs confines them to the ground surface rather than the drivable road. Recent work has begun adapting geometric ground segmentation to solid-state sensors, though only for hardware acceleration and for ground alone~\cite{ss_ground_accel}.

\subsection{Learned Point-Cloud Semantic Segmentation}

Data-driven semantic segmentation, in contrast, can label ground and road either separately or jointly, and it has spread rapidly within the autonomous driving community thanks to its ability to adapt to different sensors and scenarios. In these methods, a neural network assigns a class to every point from previously trained weights~\cite{lidar_semseg_review,seg_survey}. Projection-based methods first convert the cloud into a range image and apply 2-D convolutions for real-time inference, as in RangeNet++~\cite{rangenet} and SalsaNext~\cite{salsanext}. Point- and voxel-based networks instead operate on the 3-D structure directly: RandLA-Net~\cite{randlanet} samples large clouds efficiently, and Cylinder3D~\cite{cylinder3d} applies sparse convolutions over a cylindrical partition. More recent architectures move beyond sparse convolution: WaffleIron~\cite{waffleiron} shows that competitive accuracy is achievable with a backbone built almost entirely from multilayer perceptrons and dense 2-D convolutions, while Point Transformer V3~\cite{ptv3} uses attention on large point clouds and defines the current state of the art on major outdoor benchmarks. A more recent transformer, RWAFormer~\cite{rwaformer}, keeps this attention-based design but specializes it for road segmentation, yielding a lighter model. Some learned models specialize in ground estimation alone: GndNet~\cite{gndnet} regresses a dense ground-elevation grid with a pillar-based network and labels points against it in real time. Despite their accuracy and dominance over the field, these networks share three costs that are very relevant on a production vehicle: they require a GPU, their decisions are hard to trace or certify~\cite{ai_safety_std}, and, most important here, they are trained almost exclusively on spinning-LiDAR datasets such as SemanticKITTI~\cite{semantickitti}. Applied to a hybrid solid-state sensor without retraining, they face a domain gap that our experiments quantify.

\subsection{Geometric Road and Curb Detection}

On a much smaller scale, a few methods segment the road geometrically rather than the ground, usually by detecting curbs as the physical boundary of the drivable area. Because a curb produces a small but sharp rise in height, it can be localized geometrically. Horv\'ath \emph{et al.}~\cite{urban_road_filter} detect urban road and sidewalk boundaries in real time by scanning each sensor channel for the height and angular discontinuities that mark a curb, and Zhang \emph{et al.}~\cite{curb_seg} first segment the road with a sliding-beam method and then locate the curbs within each segment. More recent geometric systems continue this line: Wei~\emph{et al.}~\cite{drivable_area} separate the drivable area on structured and unstructured roads by thresholding point-wise features (surface normals, curvature, and intensity) selected through an information-gain analysis. Such works show that geometric road segmentation is both feasible and fast. However, built for conventional spinning LiDAR, these methods rely on its ordered beam geometry, and its sparse rings resolve a low curb only close to the sensor. On a hybrid solid-state sensor this ring structure is absent, but its high density resolves the same curb far out across the frontal field, the very property this work exploits, and one otherwise unaddressed for this class of sensor.

\subsection{Public LiDAR Datasets and Benchmarks for Ground and Road Segmentation}

Beyond methods, there is also a second gap in the available data itself that explains the scarcity of ground and road segmentation solutions for hybrid solid-state sensors: the established and widespread autonomous-driving benchmarks on which LiDAR perception research is largely based (SemanticKITTI~\cite{semantickitti}, nuScenes~\cite{nuscenes}, and the Waymo Open Dataset~\cite{sun2020scalability}) provide extensive per-point semantic labels for their LiDAR data, ground and road among them, over thousands of real driving scenes. Every one of these benchmarks, however, uses spinning mechanical LiDARs, so the regularly spaced ring-and-azimuth structure they capture is precisely what a hybrid solid-state sensor lacks. Therefore, by definition, they cannot be used to develop or evaluate a method built for HSS data.

A few solid-state and hybrid solid-state datasets have recently appeared, bringing the right sensor but not the right task or focus. PandaSet~\cite{pandaset} is the one that comes closest: it pairs a spinning Pandar64 LiDAR with a forward-facing hybrid solid-state PandarGT and it provides semantic-segmentation labels, a drivable-surface class included. Its hybrid solid-state unit, however, is auxiliary to the mechanical one in the dataset, and PandaSet is a broad general-purpose semantic-segmentation dataset, with no standard split or leaderboard, rather than a benchmark dedicated to the drivable road or hybrid solid-state technology. From the remaining public HSS and SS datasets none targets ground and road segmentation in real-world driving: SimoSet~\cite{simoset} records automotive hybrid solid-state data, yet only for object detection, with no ground or road labels; S.MID~\cite{smid} provides hybrid solid-state (Livox Mid-360) semantic segmentation in industrial substations rather than on the road; and i2Nav-Robot~\cite{i2nav_robot} mounts a hybrid solid-state LiDAR (Hesai AT128) on a small mobile robot for multi-sensor navigation, but without ground or road labels. Even the most recent datasets fall short: LiDARDustX~\cite{wei2025lidardustx} includes a Seyond Falcon among its six LiDAR sensors and does have ground labels, yet it targets dusty off-road environments such as mines and sand quarries and provides no drivable-road class. Hence, none combines a hybrid solid-state sensor, real driving scenarios, a public benchmark, and ground-and-road semantic labels. The complete landscape has been captured in Table~\ref{tab:datasets}.

\begin{table*}[t]
\caption{Public LiDAR Datasets for Ground and Road Segmentation.}
\label{tab:datasets}
\centering
\footnotesize
\begin{tabular}{@{}llllccc@{}}
\hline
Dataset & Sensor(s) & Type & Environment & Benchmark & Ground & Road \\ \hline
SemanticKITTI~\cite{semantickitti} & Velodyne HDL-64E & Mechanical & Urban driving & \cmark & \cmark & \cmark \\
nuScenes~\cite{nuscenes} & Velodyne HDL-32E & Mechanical & Urban driving & \cmark & \cmark & \cmark \\
Waymo Open~\cite{sun2020scalability} & 1 top + 4 perimeter & Mechanical & Urban driving & \cmark & \cmark & \cmark \\
PandaSet~\cite{pandaset} & Pandar64 + PandarGT & Mechanical + Hybrid SS & Urban driving & \xmark & \cmark & \cmark \\
SimoSet~\cite{simoset} & Forward hybrid SS & Hybrid SS & Urban driving & \cmark & \xmark & \xmark \\
S.MID~\cite{smid} & Livox Mid-360 & Hybrid SS & Industrial & \cmark & \cmark & \cmark \\
i2Nav-Robot~\cite{i2nav_robot} & Hesai AT128 & Hybrid SS & Mobile robot (campus) & \xmark & \xmark & \xmark \\
LiDARDustX~\cite{wei2025lidardustx} & Falcon K1/K3 + others & Mixed & Off-road (mining) & \cmark & \cmark & \xmark \\
\textbf{HySoLiD-Drive (ours)} & \textbf{Seyond Falcon} & \textbf{Hybrid SS} & \textbf{Urban driving} & \cmark & \cmark & \cmark \\
\hline
\end{tabular}
\end{table*}

Two gaps therefore motivate this work. First, geometric segmentation of ground and road has not been addressed natively for hybrid solid-state data: existing geometric approaches either stop at the ground or assume the ring-and-azimuth structure that these sensors lack, whereas the learned alternatives are heavy, opaque, and trained on other sensor types. Second, as captured in Table~\ref{tab:datasets}, no public benchmark dedicated to ground and drivable-road segmentation in real-world driving scenarios is centered around an automotive hybrid solid-state LiDAR. The established and renowned driving benchmarks all use mechanical LiDARs as their only or primary (in the case of PandaSet) sensor, and the existing datasets centered on hybrid solid-state are either detection-only, off-road, or not focused on cars but on compact mobile robots or industrial applications. This work closes both gaps at once with GHySSeg, the first native geometric ground-and-road segmentation method for hybrid solid-state LiDAR, and HySoLiD-Drive, to our knowledge, the first public benchmark dedicated to ground and road segmentation on an automotive hybrid solid-state LiDAR in real-world driving.

\section{Method}
\label{sec:method}

\subsection{Exploiting Hybrid Solid-State LiDAR Data: From Rings to Curves}
\label{sec:method-data}

The proposed method segments each point of a raw hybrid solid-state LiDAR point cloud, in real time and by pure geometry, into ground, drivable road, and obstacles, exploiting the sensor's ray distribution, front-facing nature, and high 3-D density. That density is the key enabler: where a spinning LiDAR, with its sparse and widely spaced rings, may confuse a curb with sensor noise or with the road's natural camber, an HSS sensor places dozens of points on either side of that curb (and often on its very face) so that even a centimetric rise in height remains identifiable at long range.

Hybrid solid-state LiDARs are not entirely free of moving parts. They steer their beams in several ways, using micro-electro-mechanical system (MEMS) mirrors, rotating prisms, or polygonal and galvanometer mirrors, among others~\cite{lidar_ad_survey,solid_state}. Our method does not assume any particular steering technology; it relies only on two properties that these designs share: a high point density concentrated in a front-facing field of view, and a scan pattern that departs from sparse, cleanly spaced $360^\circ$ rings. That scan pattern gives a loose notion of order: successive sweeps trace concave curves, or pseudo-rings, which may be irregular or even intersecting, yet, being produced by a deterministic physical scan, form a repeatable structure that the method can traverse.

\begin{figure}[t]
  \centering
  \includegraphics[width=\columnwidth]{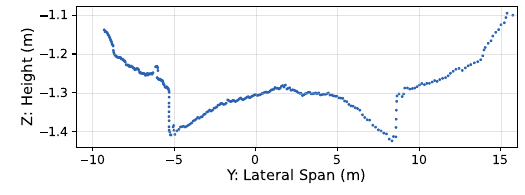}
  \caption{Height profile along a single HSS LiDAR scan curve crossing a four-lane road.}
  \label{fig:zjump}
\end{figure}

Seen from above (Fig.~\ref{fig:pc-falcon}), the curves look disordered near the sensor: early curves cross and mix at the center of the field of view, yet separate cleanly at range, each taking a concave shape. This geometry naturally yields two clear signatures of the road edge. The first is range-based: where the road ends, the curves bend back toward the sensor, because a rise in height interrupts them; the raw point cloud thus marks the road edge directly, as can be seen in Fig.~\ref{fig:pc-falcon}. The second signature is vertical. A road is physically bounded by elements that rise abruptly from the drivable surface (curbs, guardrails, fences, or walls), each ending the road with a sudden step in height that a 3-D LiDAR resolves directly. Figure~\ref{fig:zjump} plots the height profile along a single scan curve of an HSS LiDAR crossing a four-lane road. We can observe the gentle camber of the road, bounded on both sides by a sharp, easily identifiable step of roughly 0.1\,m where the curb rises suddenly.

Our method turns these signatures into a purely geometric segmentation pipeline. From the raw LiDAR point cloud $\mathcal{P}=\{\mathbf{p}_i\}_{i=1}^{N}$ (with $\mathbf{p}_i=(x_i,y_i,z_i)$), organized into the ordered scan curves $\{S_j\}$ just described, GHySSeg computes a per-point labeling
\begin{equation}
\ell = f\!(\mathcal{P}),\qquad f:\mathbb{R}^{N\times 3}\to\mathcal{L}^{N},
\label{eq:problem}
\end{equation}
assigning each point one of three nested labels, $\mathcal{L}=\{\text{non-ground},\text{ground},\text{road}\}$ with $\text{road}\subseteq\text{ground}$, by pure geometry from a single frame.

Both ground and road segmentation share the same per-curve traversal, processing each individually traversable scan curve and reasoning about the height along it. What separates them is the nature of what each recovers. The ground does not need to be smooth or bounded, so it can be segmented coarsely, by a tolerant region growing outward from the vehicle. The road, in contrast, is a smooth surface with a clearly defined boundary, so it must be delimited finely, following each curve out to the edge its signatures reveal. Coarse or fine, both stages are single-shot: each segments the current scan alone, without temporal accumulation or tracking, so a complete result is available from the very first frame.

\begin{figure*}[t]
  \centering
  \includegraphics[width=\textwidth]{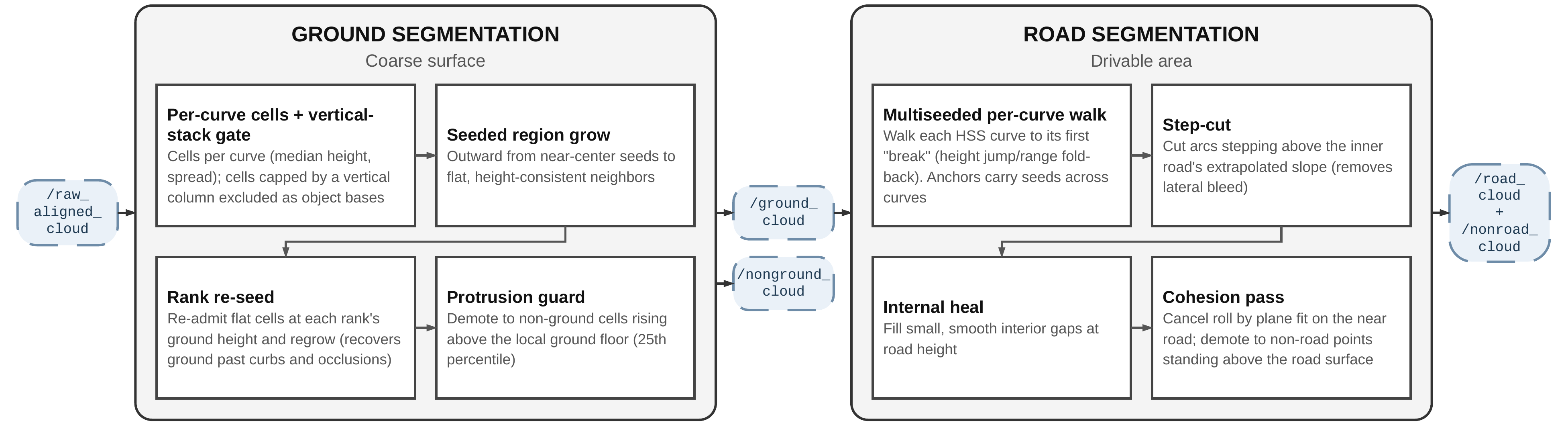}
  \caption{Overview of the proposed GHySSeg pipeline, from the raw aligned point cloud to the three per-point labels.}
  \label{fig:pipeline}
\end{figure*}

Figure~\ref{fig:pipeline} summarizes the complete proposed method. From the raw, aligned point cloud, ground segmentation forms per-curve cells, seeds the near central curves, and grows a flat, height-consistent surface outward in range order; only the points that lie inside this surface are considered ground. These ground points feed road segmentation, which walks each curve outward from the vehicle centerline to the first boundary, propagates seeds across curves, and closes with a cohesion pass that enforces spatial consistency. The result is three per-point labels for the complete point cloud, non-ground (obstacles), ground, and road. These are ready to be consumed by downstream autonomous-driving modules such as occupancy-grid maps, obstacle detection, or navigation. Each stage of the method is described in depth below.

The complete method has been implemented in C++ and wrapped as ROS~2 (Humble) nodes. Therefore, the pipeline can be integrated directly into a real autonomous-driving stack: it subscribes to the point cloud streamed by the LiDAR's driver and publishes each label on its own standard PointCloud2 topic, ready for the downstream modules to consume. In this form the method runs online, at the sensor's full frame rate, and on an on-board CPU alone.

\subsection{Curve-Cell Ground Segmentation}
\label{sec:method-ground}

Ground segmentation opens the proposed segmentation pipeline (Fig.~\ref{fig:pipeline}): it produces the coarse ground surface on which the fine road stage is later run. Its design exploits the intrinsic characteristics of hybrid solid-state data. Because the sensor concentrates a very high density into a front-facing FOV, each small patch of ground carries many measurements, so a per-patch median height and spread are stable and well-supported statistics, and the tightly spaced curves (unlike spaced rings) make neighboring patches genuinely adjacent. This density is what lets the method work by pure local geometry, and it shapes two design choices at the core of the method. First, it is fully relative: it fits no global or per-zone plane and imposes no absolute height threshold, deciding ground purely from local neighbors, so it respects the road's camber, its slopes and ramps, and the roll the sensor picks up in turns. Second, it reasons vertically as well as along the surface. Because the dense, front-facing cloud places many points on every object's profile per frame, the method can recognize an obstacle's flat base from the column of returns standing directly above it and keep it out of the ground. These design choices, when combined, give the method its robustness and precision, respectively.

\begin{figure}[t]
  \centering
  \includegraphics[width=\columnwidth]{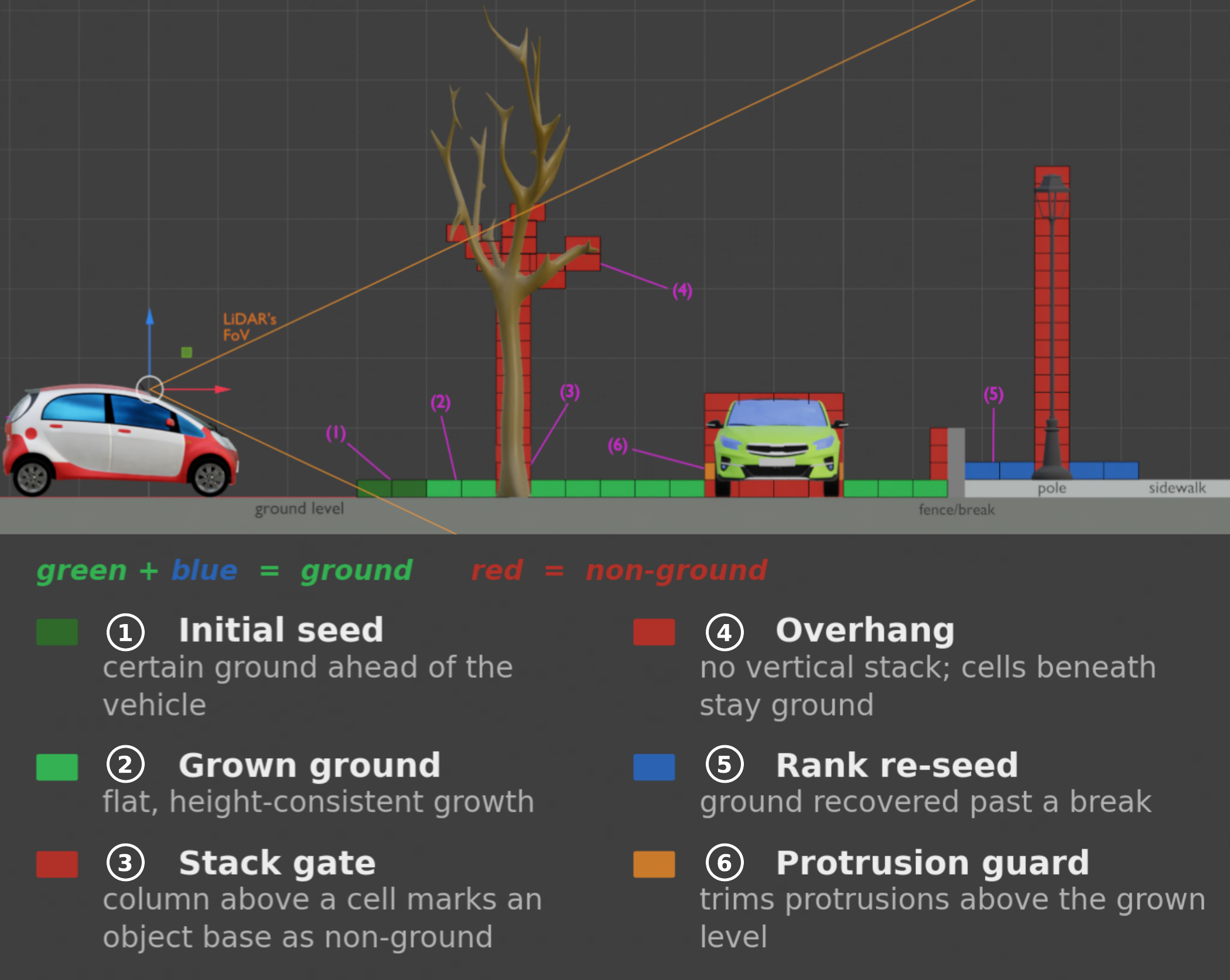}
  \caption{Our ground segmentation method on a schematic scene.}
  \label{fig:ground-pipeline}
\end{figure}

The method is based on per-curve cells and can be visualized step-by-step in Fig.~\ref{fig:ground-pipeline}. Each individual scan curve, already indexed by the sensor driver, is split into small cells of roughly fixed point count, whose width is allowed to grow with range to adapt to distant, sparser curves. Each cell $c$ of $n$ points is summarized by three descriptors it carries: its planar centroid $(x,y)_c$, its median height $\bar{z}_c$, and a trimmed vertical spread $\sigma_c$,
\begin{equation}
\bar{z}_c = \operatorname{median}\{z_{\mathbf{p}} : \mathbf{p}\in c\},\qquad \sigma_c = z_{(n-2)}-z_{(3)},
\label{eq:descriptors}
\end{equation}
where $z_{(1)}\le\dots\le z_{(n)}$ are the cell's sorted point heights, so the trimmed height spread $\sigma_c$ drops the two highest and two lowest returns. The curves are ranked from near to far, and the neighbors of a cell are its lateral neighbors on the same curve together with the cells of the rank-adjacent curves within a radial reach that widens with range, as the curves spread apart. The surface is then grown from seeds: the frontal cells immediately ahead of the vehicle centerline are marked ground~\sfig{1}, since this area is always ground. From each seeded ground cell the region expands to a neighbor $c'$ whenever $c'$ is flat and height-consistent with its parent cell $c$~\sfig{2},
\begin{equation}
\sigma_{c'} \le \tau_{\mathrm{flat}}(r,|y|)\quad\text{and}\quad |\bar{z}_{c'}-\bar{z}_{c}| \le \tau_{\mathrm{level}}(r,|y|),
\label{eq:grow}
\end{equation}
and is not flagged as a stack base~\eqref{eq:stack}. Each tolerance is range-adaptive, $\tau(r,|y|)=\tau_{0}+s_{r}\,r+s_{y}\,|y|$, relaxing with range $r$ and lateral offset $|y|$ to match the thinning data of the sensor. The outcome is a single connected surface, locally flat and locally level, grown outward from beneath the vehicle by local comparison alone.

This growth, however, is blind to vertical structure: where a curve crosses a tree trunk, pole, or car flank, it meets a flat, ground-height cell, so the region climbs onto the object and every obstacle gains a false ground ``skirt.'' No flatness or height threshold can remove it without also rejecting real curbs, camber, and ramps. The key lies on the one dimension the growth ignores, the vertical axis: true ground has open space above it, whereas an object base sits beneath a rising column of cells. Before seeding, our method therefore tests each cell $c$ against those above it. Writing $z_{\mathrm{col}}(c)$ for the top of the continuous column of cells above $c$ within a radius $R_{\mathrm{stack}}$ (a column that climbs only while consecutive height gaps stay below $\Delta_{\mathrm{gap}}$), the cell is classified as
\begin{equation}
c \text{ is }
\begin{cases}
\text{an object base}, & z_{\mathrm{col}}(c)-\bar{z}_c \ge h_{\mathrm{stack}},\\
\text{ground (clearance)}, & \text{otherwise}.
\end{cases}
\label{eq:stack}
\end{equation}
An unbroken stack rising past $h_{\mathrm{stack}}$ marks the cell as an object~\sfig{3}, while a gap wider than $\Delta_{\mathrm{gap}}$ before that height reads as clearance, keeping ground under overhangs and bridges~\sfig{4}. The sensor's dense sampling is what makes this reliable: a protruding object returns many points at successive heights, forming a solid column of cells rather than the few scattered returns a sparse scan would leave. Marked cells are simply skipped during seeding and growth, halting the surface at each object's base at the source; the ablation in Section~\ref{sec:res-ablation} attributes most of the method's precision to this gate.

Two additional passes finish and refine the surface. A curb, median, or fence rightly stops the ground growth, yet ground often continues beyond such a break, out of the reach of a single connected region; we therefore re-seed by range rank, taking the median height $\hat{h}_j$ of the ground already grown in each curve $S_j$ and re-admitting any flat, non-object cell whose height matches it within a tolerance,
\begin{equation}
\begin{aligned}
\hat{h}_j &= \operatorname{median}\{\bar{z}_c : c\in\text{ground}\cap S_j\},\\
|\bar{z}_c-\hat{h}_j| &\le \tau_{\mathrm{reseed}},
\end{aligned}
\label{eq:reseed}
\end{equation}
then growing again from these seeds to reclaim falsely flagged non-ground after a discontinuity~\sfig{5}. Because admission demands both a height match and further growth, obstacles, being non-flat or above ground height, stay out. Finally, a protrusion guard loop demotes to non-ground any surviving cell that rises more than a small margin above its local ground floor,
\begin{equation}
\bar{z}_c - \varphi_c > \tau_{\mathrm{excess}} \;\Rightarrow\; \text{demote } c,
\label{eq:protrusion}
\end{equation}
where the floor $\varphi_c=Q_{25}\{\bar{z}_{c'} : c'\in\text{ground},\, \|(x,y)_{c'}-(x,y)_c\|\le R_{\mathrm{prot}}\}$ is estimated as a low (25th) percentile of the surrounding ground heights, clearing residual low clutter such as vehicle wheels or wall lips~\sfig{6}.

The result is a coarse but robust and precise ground surface across the complete frontal FOV of the HSS LiDAR sensor, computed by pure geometry in a few milliseconds. This surface (the segmented points labeled ground) is what the proposed road segmenter refines in order to extract only the drivable area.

\subsection{Per-Curve Road Segmentation}
\label{sec:method-road}

The proposed ground segmentation method is too coarse to segment the road: cell-based approaches cannot distinguish the ego road from the flush or raised surfaces around it (sidewalks, parking strips, adjacent roads), which are just as flat and continuous. Telling them apart means resolving the centimetric curb (at worst) that bounds the road without mistaking a slope or a cambered road for one. This is difficult on sparse data, which is why widely used geometric ground filters such as Patchwork++~\cite{patchworkpp} and Autoware's ray-ground filter~\cite{autoware} stop at ground segmentation, and why the few geometric methods that reach the road at all are close-range and rely on the ring structure of a spinning sensor (Section~\ref{sec:related}).

A hybrid solid-state point cloud lifts both limitations. Its density resolves a curb as a clear step in height even at range, and its concave, non-ring curves add a second, purely geometric and robust signature. A road edge therefore announces itself twice over in an HSS point cloud: a step in height as the curb (or fence, guardrail, or wall) rises, and a fold-back in range as the curve bends inward where the boundary cuts it, creating a visible seam between the road and the rest of the point cloud, as shown in Fig.~\ref{fig:pc-falcon}. Both cues are local and relative, so they survive camber, bank, grade, and the roll the sensor experiences in turns.

\begin{figure}[t]
  \centering
  \includegraphics[width=\columnwidth]{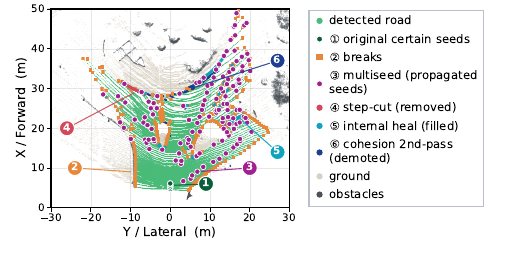}
  \caption{Internal process of the proposed road segmenter on a real frame at the entry to a roundabout with an exit on the right.}
  \label{fig:road-pipeline}
\end{figure}

Figure~\ref{fig:road-pipeline} displays the complete road segmentation process applied to a real hybrid solid-state point cloud, with the intermediate states and inner mechanisms labeled~\sfig{1}--\sfig{6}.

The heart of the proposed method is a per-curve walk. As in the ground stage, the points immediately ahead of the vehicle's center are seeded as road~\sfig{1}, since the vehicle is, by definition, driving on it. The walk then proceeds outward along each individual curve point by point (index $k$), to the left and to the right, until it meets the first \emph{break}~\sfig{2}: a step in height between $k$ and a point $k_w$ one short window ahead,
\begin{equation}
|z_{k_w}-z_k| \ge \tau_{\mathrm{step}}(d),\qquad d=\|(x,y)_{k_w}-(x,y)_k\|,
\label{eq:sig-step}
\end{equation}
or a fold-back in range, where the forward coordinate $x$ turns back toward the sensor and, about one arc window $A$ ahead (at $k^{+}$), drops below the road's local tangent,
\begin{equation}
\begin{aligned}
x_{k_w}-x_k &\le -\,\tau_{\mathrm{bend}}(r_k), \quad\text{and}\\
x_{k^{+}} &\le \hat{x}_{\mathrm{road}}(k^{+})-\tau_{\mathrm{curv}},
\end{aligned}
\label{eq:sig-bend}
\end{equation}
with the tangent $\hat{x}_{\mathrm{road}}(k^{+})=x_k+(x_k-x_{k^{-}})\,A_{\mathrm{ahead}}/A_{\mathrm{behind}}$ linearly extrapolated from a point $k^{-}$ one arc window behind $k$, where $A_{\mathrm{ahead}}$ and $A_{\mathrm{behind}}$ are the forward and backward arc lengths. The segmented road on that curve is then the arc between its two breaks $k_L$ and $k_R$,
\begin{equation}
\mathrm{road}(S_j)=\{\mathbf{p}_k\in S_j : k_L<k<k_R\},
\label{eq:road-arc}
\end{equation}
so the boundary follows the road as it turns instead of being clipped to a fixed lateral band, allowing it to adjust to tight curves or forks. This tangent term is what keeps the range-break condition honest: a concave curve already dips toward the sensor at its own center (its global range minimum), which would create a spurious road edge, so a fold-back is accepted only where the curve also drops below the road's local tangent, as a real curb or road edge does, and the curve's natural apex does not. Every threshold grows with distance except the fixed curb depth $\tau_{\mathrm{curv}}$, which keeps this test range-robust.

On a straight road, seeding every curve at its lateral center ($y=0$) would suffice. However, real roads bend, fork, and merge, so the road is carried outward curve by curve. Processed from near to far, each road region already found seeds the next: instead of a single center point it drops several anchors~\sfig{3} spread across its width, so forks and widening lanes are all followed and no single seed can drift or collapse the growth. Each anchor $\mathbf{a}$ begins a walk on the nearest point $\mathbf{p}$ of the following curve, kept only if its local height matches that specific anchor's,
\begin{equation}
|z_{\mathbf{p}}-z_{\mathbf{a}}| \le \tau_{\mathrm{seed}},
\label{eq:multiseed}
\end{equation}
and if the resulting arc hugs the previous region in range, width, and relative height. These local, relative gates reject a walk that has climbed onto a raised sidewalk without ever penalizing a legitimately banked or graded road.

Additionally, three guard mechanisms strengthen the walk where it is most vulnerable to errors (road segments without a curb, flush sidewalks, banked turns, strong body roll, and stray early returns).

\emph{Step-cut}~\sfig{4} removes lateral bleed. If the walk slips past a missing or buried curb, it continues onto the sidewalk and may wrongly seed the next curve there, creating a cascading false-positive error. Each curve's segmented road is therefore checked outward against the slope of the inner road extrapolated across the gap: a genuinely banked road continues that slope and is kept, whereas a sidewalk sits a step above it and is cut, along with everything beyond.

\emph{Internal heal}~\sfig{5} removes false-negative holes. Interior gaps that are flanked by road on both sides, short, smooth, and at road height are filled in, closing the phantom non-road specks (caused by stray early returns) that a downstream module could mistake for an obstacle ahead.

\emph{Cohesion pass}~\sfig{6} removes raised bleed from the road the walk has already traced. As a final step the cloud is leveled for this pass alone using a robust plane fit on the near road, which cancels the roll the vehicle induces while turning. A road-height reference surface is then grown outward from the trusted near road, stopped wherever it turns rough or steps up so that it never climbs a sidewalk; any road point standing more than a small margin above that reference is demoted as raised bleed, and only when its neighbors are demoted too, so no isolated holes are opened.

Algorithm~\ref{alg:road} compresses the complete road segmentation method, from the per-curve walk and cross-curve propagation to the described guards.

\begin{algorithm}[t]
\caption{Per-Curve Road Segmentation}
\label{alg:road}
\begin{algorithmic}[1]
\REQUIRE ground-segmented cloud organized into scan curves $S_1,\dots,S_M$ (near-to-far)
\ENSURE per-point road / non-road labels
\STATE seed as road the points near the vehicle center ($|y|\le w_{\mathrm{seed}}$) on the near curves
\FOR{$j=1$ to $M$}
\FOR{each road seed on curve $S_j$}
\STATE $k_L,k_R \gets$ first break outward on each side (height step~\eqref{eq:sig-step} or fold-back~\eqref{eq:sig-bend})
\STATE label the arc $k_L$--$k_R$ of $S_j$ as road
\ENDFOR
\STATE place $N_a$ anchors evenly across the width of $\mathrm{road}(S_j)$
\FOR{each anchor $\mathbf{a}$}
\STATE seed nearest $\mathbf{p}$ on $S_{j+1}$ if $|z_{\mathbf{p}}-z_{\mathbf{a}}|\le\tau_{\mathrm{seed}}$ and the arc stays consistent
\ENDFOR
\ENDFOR
\STATE \textit{\# refinement passes (guards):}
\STATE \textbf{if} a segment steps $>\tau_{\mathrm{bleed}}$ above the inner-road slope \textbf{then} demote it and all beyond \quad(step-cut)
\STATE \textbf{if} a non-road point is flanked by road at road height \textbf{then} relabel it road \quad(internal heal)
\STATE \textbf{if} a road point rises $>\tau_{\mathrm{coh}}$ above the roll-leveled near-road reference (with neighbors) \textbf{then} demote it \quad(cohesion pass)
\RETURN road labels
\end{algorithmic}
\end{algorithm}

The proposed road segmentation method is thus purely geometric and single-shot, and because it reasons one curve at a time, essentially in one dimension, it runs in real time on a single CPU core. It returns the ego road, the road the vehicle is driving on, cleanly separated from the flat non-drivable ground around it, and holds up on the hardest scenarios (roundabouts and sharp turns included), as shown in Section~\ref{sec:res-ablation}.

\section{Experimental Setup}
\label{sec:setup}

As Section~\ref{sec:related} established, the few hybrid or solid-state-centric datasets that exist target object detection or non-automotive scenes, and none is a benchmark dedicated to per-point ground and road segmentation for an automotive hybrid solid-state sensor in real-world driving scenarios. Therefore, developing and evaluating the proposed method has required recording and labeling our own data. This section describes the platforms and sensor that have made this work possible, the resulting publicly released HySoLiD-Drive benchmark, and the experimental protocol followed to keep the evaluation and comparison fair.

\subsection{Platforms and Sensor}
\label{sec:setup-platforms}

The method has been developed and evaluated on real driving data from two research autonomous vehicles at Universidad Carlos III de Madrid: ATLAS, a retrofitted first-generation electric Mitsubishi i-MiEV~\cite{atlas}, and Kalipso, a modified Hyundai Ioniq plug-in hybrid. Two vehicles have been used instead of one because their different geometry forces the sensor to sit at a different height and pitch, verifying that the geometric method generalizes rather than being tuned to a single vehicle and mount.

Both carry the same Innovusion (now Seyond) Falcon Kinetic, a long-range hybrid solid-state LiDAR~\cite{falcon_kinetic}. It steers a single 1550\,nm fiber laser with a rotating polygon mirror and an oscillating galvanometer, covering a $120^\circ \times 25^\circ$ field of view with an angular resolution of $0.18^\circ \times 0.24^\circ$ that densifies to $0.09^\circ \times 0.08^\circ$ inside the sensor's region of interest. It produces 1500 scan lines per second with a range accuracy of $\pm2$\,cm and a maximum range of 250\,m at 10\% reflectivity; a typical frame contains approximately 88\,000 points.

On both vehicles the sensor is mounted front-facing at the top of the windshield. This is the standard position for this sensor class, also adopted in production integrations such as the Xiaomi SU7, because the vehicle's own hood absorbs part of the roughly 2\,m near-field blind zone inherent to the sensor.

The single-sensor scope is deliberate: a production vehicle carries only one HSS LiDAR, and HySoLiD-Drive is built for exactly that setting. GHySSeg, moreover, exploits only the two properties every hybrid solid-state sensor shares, a dense front-facing field of view and a non-ring, per-curve scan pattern (Section~\ref{sec:method-data}), so it is not Seyond- or Falcon-exclusive. Furthermore, the two-vehicle split already probes the method's generality along a key axis of variation for a geometric method, the sensor pose (Section~\ref{sec:setup-baselines}). Extending the same transfer to other sensors is left to future work.

Point clouds are produced at the sensor's native 10\,Hz and decoded in real time by the manufacturer's ROS~2 driver from Ethernet packets into PointCloud2 messages. A static calibration stage compensates for the constant mounting pitch and roll. The compensated point cloud is the only input to the proposed method (see Fig.~\ref{fig:pipeline}). 

All experiments, for all methods, ours and the state-of-the-art baselines', have been run on a single portable computer with an AMD Ryzen 7 7435HS CPU, 24\,GB DDR5 4800\,MT/s RAM, and an RTX~4060 8\,GB GPU, so that reported runtimes and performance are directly comparable.

\subsection{The HySoLiD-Drive Benchmark}
\label{sec:setup-dataset}

Complete driving routes have been recorded with both vehicles in Leganés, Madrid, under normal traffic, with a human driver covering the full range of everyday driving scenarios: urban streets, motorways, roundabouts, sharp turns, forks, ramps, elevated roads, crosswalks, parking entries and exits, and roads of one to four lanes. These drives have yielded more than 7000 sequential raw frames, which have been the basis for developing the method but carry no ground truth.

From them, 100 frames have been handpicked for scenario diversity, evenly split between the two vehicles (50 ATLAS, 50 Kalipso), and labeled manually point by point into four nested classes: non-ground, ground, road, and ego-road. The nesting allows a single annotation to serve both tasks: every ego-road point is a road point and every road point is a ground point. Ground segmentation is evaluated against the ground class, and road segmentation against the ego-road class, which is what the method's road output denotes. The even split allows accuracy to be reported per vehicle throughout.

The dataset was originally intended only for the experiments and results of the presented work, but as labeled hybrid solid-state data is scarce and per-point 3-D annotation is very costly we have decided to release it instead as HySoLiD-Drive: to our knowledge the first public benchmark targeting ground and road segmentation on an automotive hybrid solid-state LiDAR. The benchmark data, both the labeled frames and the unlabeled complete sequences, is archived on Zenodo under a permanent DOI~\cite{hysolid_drive_dataset}. The accompanying public repository~\cite{hysolid_drive_code} provides the data-loading and evaluation tools, the documentation, the submission rules, and a leaderboard seeded with the results of this work and open to community submission.

Beyond the results reported in Section~\ref{sec:results}, the benchmark is intended as a resource for the community. The labeled split supports quantitative evaluation of any ground or road segmentation method, geometric or learned, on hybrid solid-state data under real driving conditions. Moreover, the two unlabeled complete sequences (3500 frames per vehicle) support self-supervised learning, domain adaptation from mechanical to hybrid solid-state data, runtime profiling under continuous driving, temporal and multi-frame approaches, and the development of new geometric methods. Across both splits, the even division between two vehicles with different mounting geometries makes generalization across sensor pose measurable. It is modest in size next to SemanticKITTI, but its scenario diversity and, above all, its sensor make it a starting point for work that established mechanical-oriented datasets cannot support: the automotive industry is moving toward solid-state sensing, yet every major public benchmark remains focused on mechanical LiDAR, so methods for the sensors now entering production cannot be developed or validated on public data. HySoLiD-Drive is a first step toward closing that gap.

\subsection{Baselines and Protocol}
\label{sec:setup-baselines}

The proposed method has been compared against six modern 3-D LiDAR segmentation methods, three per task, all of them reviewed in Section~\ref{sec:related}. They have been selected to represent the current state of the art and to span both schools of thought identified there (Table~\ref{tab:accuracy}).

The selection is deliberately asymmetric, reflecting where each school is strongest. Geometric methods dominate ground estimation, where the physical flatness of the surface is a reliable cue, so two of the three ground baselines are geometric (Patchwork++ and CSF) and one learned (GndNet). Learned methods dominate strict road segmentation, where the boundary is more semantic than geometric, so two of the three road baselines are learned (WaffleIron and PTv3). The one geometric road baseline is Patchwork++ reused: since no publicly available geometric road segmenter is free of a spinning LiDAR's ring structure (Section~\ref{sec:related}), we have tuned the dominant geometric ground filter so strictly on the labeled frames that only the smooth ego-road surface survives and its ground output then serves as a road estimate.

All six are designed for or trained on spinning mechanical LiDAR data, since public benchmarks provide only such data and mechanical scanning remains the established technology. The learned baselines have therefore been evaluated without retraining: no public labeled hybrid solid-state dataset for ground and road segmentation of sufficient size exists for training. The geometric baselines, by contrast, have been tuned using their available parameters and are reported at their best configuration on our data.

The proposed method has been developed and tuned on Kalipso alone. To avoid rewarding that tuning, its parameters have been frozen and applied unchanged to ATLAS; ATLAS therefore acts as a held-out vehicle to test generalization and avoid overfitting. Our method is thus evaluated under a stricter protocol than the baselines, which are tuned on the full set.

With the tuning and comparison rules in place, the correctness and efficiency of every method are quantified as follows. Accuracy is reported with the metrics standard in the segmentation literature: intersection over union (IoU) as the primary score, together with precision ($P$), recall ($R$), and F1 score, which localize the source of each error,
\begin{equation}
\begin{aligned}
&P = \frac{\mathrm{TP}}{\mathrm{TP}+\mathrm{FP}}, \quad R = \frac{\mathrm{TP}}{\mathrm{TP}+\mathrm{FN}},\\
&\mathrm{F1} = \frac{2\,\mathrm{TP}}{2\,\mathrm{TP}+\mathrm{FP}+\mathrm{FN}}, \quad \mathrm{IoU} = \frac{\mathrm{TP}}{\mathrm{TP}+\mathrm{FP}+\mathrm{FN}},
\end{aligned}
\label{eq:metrics}
\end{equation}
where $\mathrm{TP}$, $\mathrm{FP}$, and $\mathrm{FN}$ count the true positive, false positive, and false negative points for the evaluated class (ground or ego-road); all metrics are broken down per vehicle. Real-time performance is measured on the 3500 continuous unlabeled frames of each vehicle, timing computation only (disk and message I/O excluded), after discarding 10 warm-up frames, with the same frame count for every method.

\section{Results}
\label{sec:results}

The proposed method has been thoroughly evaluated through four experiments, each addressing a different property: its segmentation accuracy against the state of the art (Section~\ref{sec:accuracy}), its real-time performance under real-world driving conditions (Section~\ref{sec:res-timing}), the contribution of each ideated mechanism, isolated in an ablation study (Section~\ref{sec:res-ablation}), and its behavior and limitations in the hardest scenes, examined qualitatively (Section~\ref{sec:res-qual}). Every experiment has been run on HySoLiD-Drive, so all results are reproducible on the public benchmark.

\subsection{Segmentation Accuracy}
\label{sec:accuracy}

Table~\ref{tab:accuracy} is the product of running the proposed ground and road segmentation pipeline and the state-of-the-art methods on the 100 distinct, labeled frames from HySoLiD-Drive. Results are split by vehicle (Kalipso, on which the method has been tuned, and ATLAS, on which it runs unchanged) and over the full set. For each metric, the best result is shown in bold and the second best underlined.

\begin{table*}[t]
\caption{Ground and Road Segmentation Accuracy on HySoLiD-Drive Benchmark.}
\label{tab:accuracy}
\centering
\footnotesize
\begin{tabular}{@{}llcccccccccccc@{}}
\hline
 & & \multicolumn{4}{c}{ATLAS ($N{=}50$)} & \multicolumn{4}{c}{Kalipso ($N{=}50$)} & \multicolumn{4}{c}{All ($N{=}100$)} \\
\cline{3-6}\cline{7-10}\cline{11-14}
Method & Type & IoU & P & R & F1 & IoU & P & R & F1 & IoU & P & R & F1 \\ \hline
\multicolumn{14}{@{}l}{\emph{Ground segmentation}} \\
GHySSeg (ours) & Geometric & \textbf{0.882} & \textbf{0.963} & 0.914 & \textbf{0.936} & \textbf{0.911} & \textbf{0.960} & 0.947 & \textbf{0.953} & \textbf{0.896} & \textbf{0.961} & 0.931 & \textbf{0.944} \\
Patchwork++~\cite{patchworkpp} & Geometric & \underline{0.874} & \underline{0.917} & 0.948 & \underline{0.931} & \underline{0.879} & \underline{0.931} & 0.940 & \underline{0.934} & \underline{0.877} & \underline{0.924} & 0.944 & \underline{0.933} \\
CSF~\cite{csf} & Geometric & 0.837 & 0.855 & \underline{0.979} & 0.903 & 0.863 & 0.863 & \textbf{0.999} & 0.925 & 0.850 & 0.859 & \textbf{0.989} & 0.914 \\
GndNet~\cite{gndnet} & Learned & 0.747 & 0.751 & \textbf{0.993} & 0.851 & 0.724 & 0.737 & \underline{0.978} & 0.835 & 0.736 & 0.744 & \underline{0.985} & 0.843 \\
\hline
\multicolumn{14}{@{}l}{\emph{Road segmentation (ego-road)}} \\
GHySSeg (ours) & Geometric & \textbf{0.921} & \textbf{0.977} & \textbf{0.940} & \textbf{0.956} & \textbf{0.892} & \textbf{0.962} & \textbf{0.926} & \textbf{0.941} & \textbf{0.907} & \textbf{0.969} & \textbf{0.933} & \textbf{0.949} \\
WaffleIron~\cite{waffleiron} & Learned & \underline{0.823} & \underline{0.930} & 0.879 & \underline{0.899} & \underline{0.747} & \underline{0.878} & 0.834 & \underline{0.853} & \underline{0.785} & \underline{0.904} & 0.856 & \underline{0.876} \\
Patchwork++ (road-tuned)~\cite{patchworkpp} & Geometric & 0.749 & 0.813 & \underline{0.903} & 0.851 & 0.705 & 0.772 & \underline{0.888} & 0.821 & 0.727 & 0.792 & \underline{0.895} & 0.836 \\
PTv3~\cite{ptv3} & Learned & 0.421 & 0.805 & 0.476 & 0.590 & 0.144 & 0.661 & 0.156 & 0.249 & 0.283 & 0.733 & 0.316 & 0.420 \\
\hline
\end{tabular}
\end{table*}

On the ground task, Table~\ref{tab:accuracy} shows that GHySSeg achieves the highest IoU (0.896) and F1 score (0.944), ahead of Patchwork++ (0.877 IoU) and CSF (0.850 IoU); GndNet, the only learned ground baseline, ranks considerably lower at 0.736. This ranking was expected: GHySSeg is designed to exploit hybrid solid-state geometry, whereas the baselines rely on generic representations or are tailored to mechanical LiDAR. Patchwork++ nonetheless remains surprisingly competitive despite its ring-based formulation, which illustrates its adaptability and explains its wide adoption for ground segmentation tasks. The decisive difference, however, is precision. Our ground method is deliberately built to maximize it, because a false ground point (a real obstacle labeled as traversable) is the safety-critical error in autonomous driving that could cause a fatal collision~\cite{yang2021introspective}. GHySSeg reaches 0.961 precision against 0.924 for Patchwork++, and is the only method that pairs high precision with high recall (0.931). CSF and GndNet achieve higher recall (0.989 and 0.985, even up to 0.999 for CSF on Kalipso), but at the cost of numerous false positives (most severely GndNet), which could let real, collidable obstacles pass as ground in a real application. These results support our thesis: on hybrid solid-state data, a native geometric method surpasses the established state of the art even where geometric methods are traditionally strongest (ground segmentation). The margin is narrower than expected, however, because mature methods such as Patchwork++ and CSF withstand the domain gap far better than learned ones, as demonstrated. Across vehicles, freezing the parameters for ATLAS costs only 0.029 IoU and 0.017 F1 score relative to Kalipso, indicating that mounting geometry and tuning have a real but minor effect on the proposed ground segmentation method.

On the road task, GHySSeg dominates every metric on both vehicles, and by a wide margin: its 0.907 IoU over the full set exceeds the runner-up, WaffleIron (0.785), by more than 0.12, while road-tuned Patchwork++ reaches only 0.727 and PTv3 collapses to 0.283. With these results we are able to identify two causes that explain this gap. First, classical purely geometric methods, even the best and most modern, struggle to refine ground into road even on dense clouds: road-tuned Patchwork++ attains the second-highest recall (0.895) but bleeds into sidewalks and non-road ground, which lowers its precision and caps its IoU. Second, the learned baselines suffer heavily from the domain gap; trained on mechanical-LiDAR datasets (KITTI, nuScenes) and applied without retraining, they face an unfamiliar topology. The effect is most pronounced for the heavy transformer approach PTv3, which stays at 0.283 IoU. WaffleIron generalizes better, holding up surprisingly well on the unfamiliar data, but still lags in recall (0.856). These results support the central claim of this article: by exploiting hybrid solid-state data characteristics, geometric road segmentation is not only feasible but comfortably outperforms non-retrained learned models and state-of-the-art non-specialized geometric methods, reaching 0.969 precision from a purely geometric formulation. The cross-vehicle comparison is equally telling: unlike on the ground task, GHySSeg scores higher on ATLAS (0.921 IoU) than on Kalipso (0.892), so no cross-vehicle penalty or re-tuning is incurred. This outcome is partly due to easier road scenes on ATLAS; even so, it is achieved with parameters tuned solely on Kalipso.

\subsection{Runtime}
\label{sec:res-timing}

\begin{figure}[t]
  \centering
  \includegraphics[width=\columnwidth]{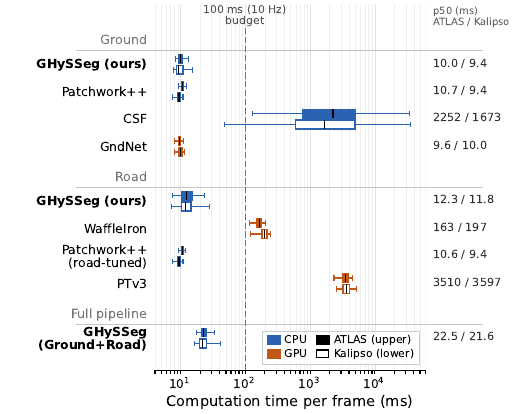}
  \caption{Per-frame computation time of every method on the continuous
    3500-frame sequences of both vehicles (logarithmic scale).}
  \label{fig:timing}
\end{figure}

The second most relevant results concern real-time performance. Real-time operation is a critical requirement for autonomous driving: an on-board perception method must deliver its output in time for the downstream navigation, decision, and control modules to act on it~\cite{ad_latency}. Our pipeline and the baseline methods have therefore been profiled on the two continuous real-world drives of HySoLiD-Drive (Section~\ref{sec:setup-baselines}) to measure and compare their processing time against a 100\,ms budget (the sensor's native 10\,Hz frame period), since exceeding it causes the method to drop frames or lag. Figure~\ref{fig:timing} shows the per-frame processing time of each method, split by vehicle, on a logarithmic scale.

The first thing that we can observe in Fig.~\ref{fig:timing} is that the complete GHySSeg pipeline (ground followed by road) runs in about 22\,ms per frame on both vehicles, which is comfortably within the 100\,ms budget; even the single worst frame (48.1\,ms on Kalipso) stays below half the frame period, leaving more than 50\,ms of headroom for downstream planning and control. Ground and road contribute similarly under nominal conditions ($\sim$10 and $\sim$12\,ms), but road is the more variable stage: its guard mechanisms activate only when needed, so a frame requiring bald-spot repair or strong roll compensation costs more than a clean straight one, whereas ground stays stable across all 7000 frames. The per-frame standard deviation remains small in both cases ($\sim$2 and $\sim$4\,ms), so latency is predictable, a property paramount for driving safety. These results demonstrate that the proposed pipeline, even with both methods stacked, achieves real-time performance running on a single CPU core.

Regarding the baselines' performance when processing hybrid solid-state data, Patchwork++ is the fastest method, marginally quicker than both GHySSeg stages at $\sim$10\,ms per frame for both tasks and vehicles with very low variance, confirming that its authors' real-time claims hold up well even on non-mechanical data. GndNet is the only other method within budget, matching Patchwork++ at $\sim$10\,ms per frame with the lowest spread of all methods; it requires a GPU to run, however, adding cost and power consumption to the vehicle. The remaining methods exceed the 100\,ms budget. CSF is the most inconsistent with a median per-frame time of 2.25\,s on ATLAS and 1.67\,s on Kalipso, and taking up to 282\,s to process a single frame; this is expected of an offline mapping and topography tool rather than a real-time segmenter. WaffleIron ($\sim$175\,ms per-frame median) and PTv3, the slowest method at $\sim$3.5\,s, both remain too heavy (despite the GPU processing) for real-time inference on a consumer-grade RTX~4060 due to their larger models and the many classes they segment beyond road.

Across vehicles, every method (not only the proposed one) shows a wider spread on Kalipso, whose more demanding route includes stretches where the sensor sees up to 200\,m ahead, and traversing curves at that range costs additional time; the mean, however, stays consistent between vehicles for GHySSeg, confirming stable performance across environments and vehicles.

\subsection{Ablation Study}
\label{sec:res-ablation}

As described in Section~\ref{sec:method}, both proposed methods share the per-curve seeding principle but pursue different goals (coarse ground versus fine road) by cumulatively stacking mechanisms, each targeting a specific failure observed in real data. Consistent with the safety priorities of autonomous driving, every mechanism is meant to raise the floor (the worst-case scenario) and favor precision, so that behavior stays predictable across scenes and catastrophic failures are avoided~\cite{yang2021introspective}. To make each stage's purpose explicit, we have performed a cumulative ablation on the 100 labeled frames of HySoLiD-Drive, grouped by scenario, that isolates every mechanism's contribution and confirms it earns its place in the reported accuracy; the results are gathered in Table~\ref{tab:ablation}.

\begin{table*}[t]
\caption{Cumulative Ablation of the Proposed Ground and Road Methods: IoU Evolution per Scenario.}
\label{tab:ablation}
\centering
\footnotesize
\begin{tabular}{@{}lccccccccccc@{}}
\hline
 & \multicolumn{8}{c}{IoU per scenario bucket} & \multicolumn{3}{c}{All ($N{=}100$)} \\
\cline{2-9}\cline{10-12}
Configuration & Crossw. & Roundab. & Sharp t. & Ramp & Intersec. & Parking & Curve & Straight & IoU & P & R \\ \hline
\multicolumn{12}{@{}l}{\emph{Ground segmentation}} \\
Seeded region grow & 0.869 & 0.852 & 0.799 & 0.732 & 0.851 & 0.851 & 0.818 & 0.808 & 0.825 & 0.860 & 0.953 \\
\quad + vertical-stack gate & 0.912 & 0.894 & 0.875 & 0.781 & 0.898 & 0.924 & 0.883 & 0.879 & 0.885 & 0.934 & 0.942 \\
\quad + rank re-seed & 0.909 & 0.891 & 0.874 & 0.776 & 0.900 & 0.922 & 0.876 & 0.878 & 0.883 & 0.922 & 0.954 \\
\quad + protrusion guard (full) & 0.910 & 0.900 & 0.892 & 0.825 & 0.916 & 0.923 & 0.878 & 0.895 & 0.896 & 0.961 & 0.931 \\
\hline
\multicolumn{12}{@{}l}{\emph{Road segmentation}} \\
Center-seeded walk & 0.839 & 0.813 & 0.633 & 0.946 & 0.922 & 0.827 & 0.953 & 0.945 & 0.882 & 0.949 & 0.924 \\
\quad + multiseed propagation & 0.797 & 0.880 & 0.739 & 0.944 & 0.926 & 0.886 & 0.933 & 0.938 & 0.900 & 0.958 & 0.937 \\
\quad + curvature veto & 0.798 & 0.881 & 0.749 & 0.940 & 0.921 & 0.884 & 0.933 & 0.939 & 0.900 & 0.949 & 0.946 \\
\quad + step-cut & 0.849 & 0.875 & 0.748 & 0.941 & 0.927 & 0.892 & 0.942 & 0.946 & 0.906 & 0.958 & 0.943 \\
\quad + internal heal & 0.842 & 0.873 & 0.745 & 0.942 & 0.925 & 0.892 & 0.942 & 0.945 & 0.905 & 0.955 & 0.945 \\
\quad + cohesion pass (full) & 0.856 & 0.889 & 0.792 & 0.940 & 0.910 & 0.898 & 0.935 & 0.937 & 0.907 & 0.969 & 0.933 \\
\hline
\end{tabular}
\end{table*}

Regarding the ground method composition, Table~\ref{tab:ablation} shows that the seeded region grow alone reaches 0.825 IoU with high recall (0.953) but modest precision (0.860), and is already stable across scenarios. The vertical-stack gate raises precision sharply, as intended, to 0.934 for only a 0.011 recall loss, improving IoU in every scenario without exception, as any vertical structure (pole, car, wall, or tree) now halts the ground growth immediately. The rank re-seed costs a negligible 0.002 IoU on average but restores recall to 0.954 by reclaiming ground beyond vertical breaks such as medians. Finally, the protrusion guard completes the method, trimming residual object skirts and gently rising surfaces to lift precision by a further 0.039, to 0.961, for a 0.023 recall trade. It benefits the ramp scenes the most (until then the hardest bucket by roughly ten points), which climb from 0.776 to 0.825 IoU, in line with the other scenarios at last.

The road segmentation chain tells a similar story. The center-seeded walk starts strong at 0.882 IoU but is markedly unbalanced: straights and mild curves reach around 0.95, whereas sharp turns fall to 0.633 and roundabouts to 0.813, because a single seed at the lateral center cannot follow a curving or branching road. Multiseed propagation equalizes this profile, trading a small drop on straights (to 0.938) for large gains where the road bends (roundabout to 0.880, sharp turn to 0.739), at the cost of making the system more prone to bleeding into non-road in difficult scenarios (crosswalk drops to 0.797). The curvature veto leaves IoU unchanged at 0.900 but shifts the precision--recall balance toward recall (from 0.937 to 0.946), suppressing the spurious break a curve's concave apex would otherwise trigger, so the walk reaches further along bending roads. The step-cut then self-corrects curves that grow past the road edge, recovering the crosswalk bucket to 0.849 and lifting the overall IoU to 0.906. The internal heal is not an accuracy mechanism (it changes IoU by only $-0.001$) but a downstream-safety fill that closes false negatives and bald spots near the sensor that would otherwise be read as obstacles and trigger spurious emergency braking. Finally, the cohesion pass completes the method, raising the hardest scenes once more (sharp turn to 0.792) and bringing both precision (0.969) and IoU (0.907) to their peaks.

The ablation thus demonstrates that each mechanism earns its place: by targeting the weakest scenarios and favoring precision, the added stages equalize performance and keep the method predictable across driving situations.

\subsection{Qualitative Results and Limitations}
\label{sec:res-qual}

\begin{figure}[t]
  \centering
  \includegraphics[width=0.45\columnwidth]{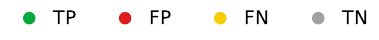}\\[2pt]
  \subfloat[Ground (Kalipso), sharp left turn: tree canopies, parked car, curbs.]{%
    \includegraphics[width=0.48\columnwidth]{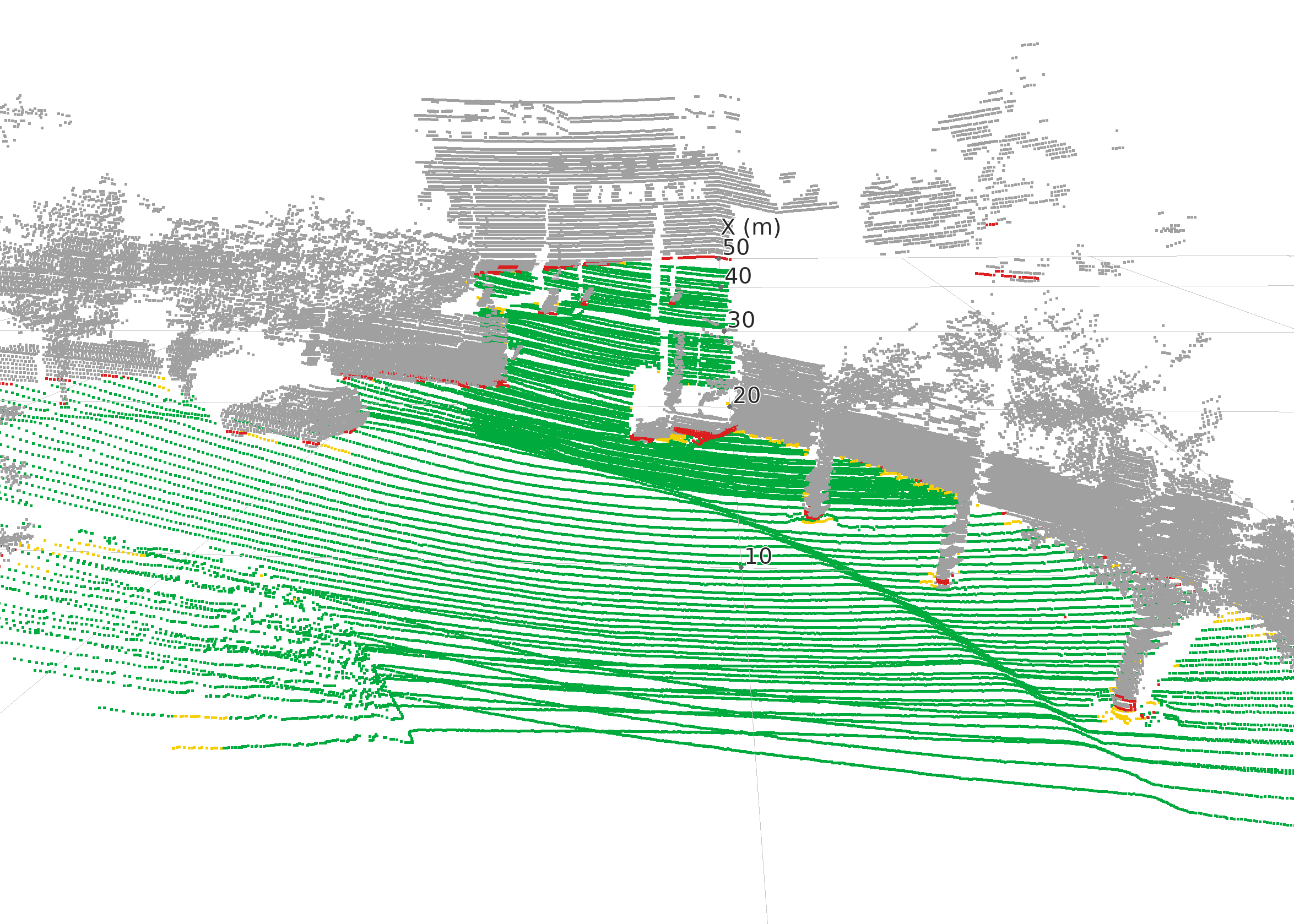}%
    \label{fig:qual-a}}
  \hfil
  \subfloat[Ground (ATLAS), vehicles ahead: ground recovered at the other lane.]{%
    \includegraphics[width=0.48\columnwidth]{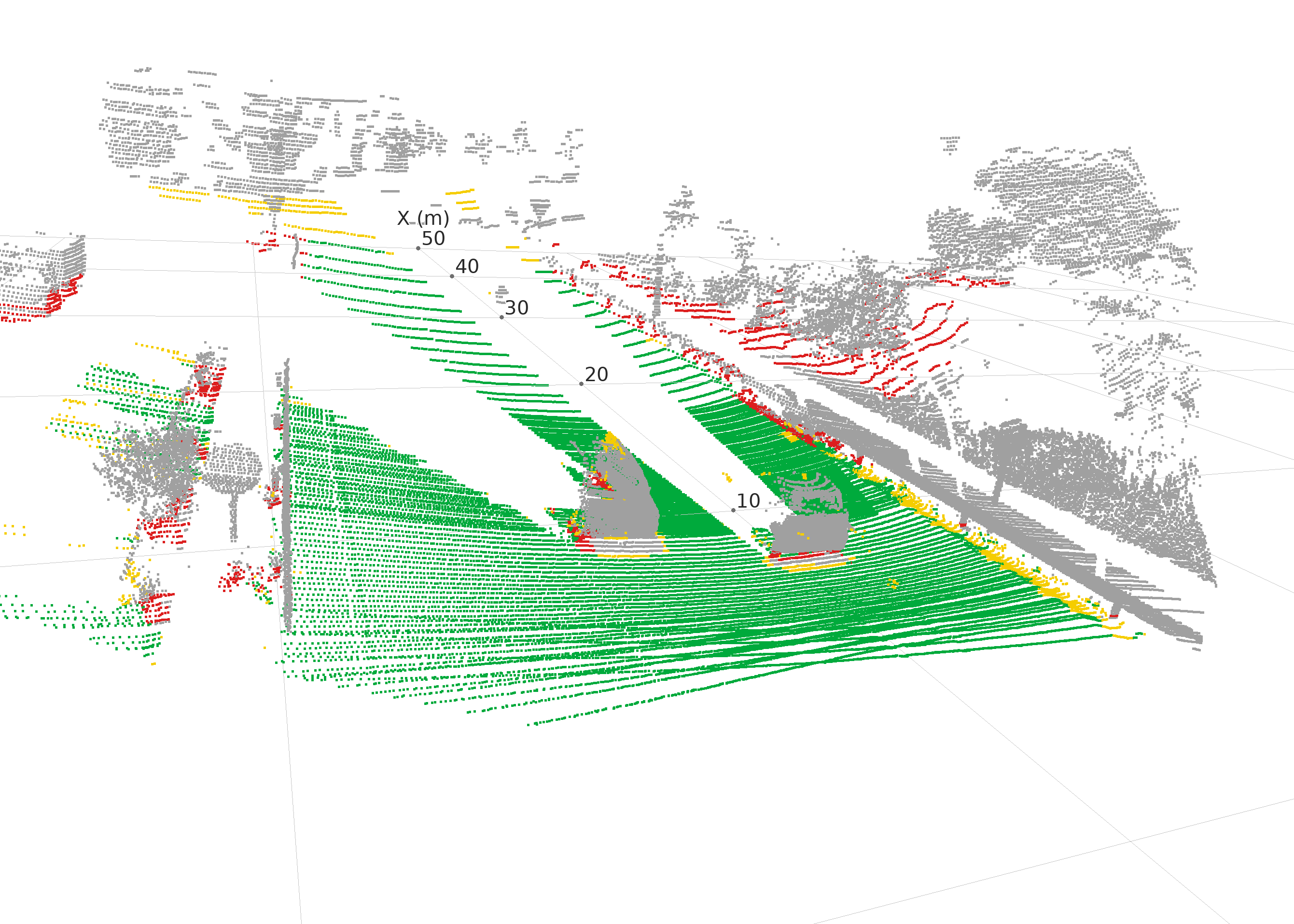}%
    \label{fig:qual-b}}\\[2pt]
  \subfloat[Road (ATLAS), right-curving urban road: far-range FN as scan lines spread.]{%
    \includegraphics[width=0.48\columnwidth]{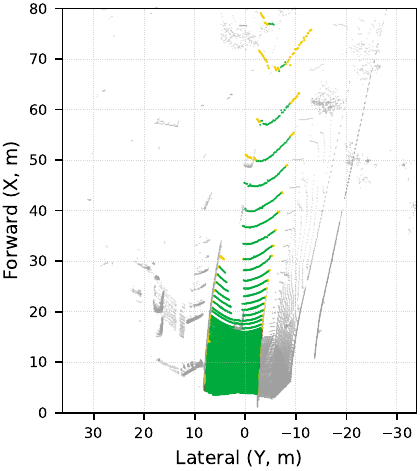}%
    \label{fig:qual-c}}
  \hfil
  \subfloat[Road (Kalipso), crosswalk: road bleed where no physical limit bounds the road.]{%
    \includegraphics[width=0.48\columnwidth]{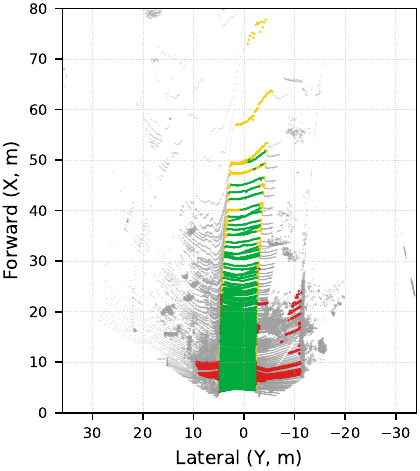}%
    \label{fig:qual-d}}
  \caption{Qualitative segmentation results on four representative frames.}
  \label{fig:qualitative}
\end{figure}

To complement the metrics with an intuitive view, Figure~\ref{fig:qualitative} presents four representative HySoLiD-Drive frames (two per task, one from each vehicle) that illustrate the strengths and known limitations of the proposed method.

Figures~\ref{fig:qual-a} and \ref{fig:qual-b} concern the ground method, shown in a 3-D perspective view so that object height and volume are visible. In Fig.~\ref{fig:qual-a}, the ego vehicle takes a sharp urban left turn past a parking entrance, a badly parked car, a fence, and leafy trees. In this scene, the vertical-stack and protrusion guards keep the ground correct almost everywhere, with false positives confined to the first one or two scan lines at the base of each rising object (tree trunks, wheels) before growth is halted. Figure~\ref{fig:qual-b} shows a motorway where the frontal view of the LiDAR is occluded by traffic. Here, our method recovers the occlusion gaps and, through re-seeding, the ground across the median on the left. This recovery is not free, however: a truck on the left and roadside vegetation on the right are wrongly seeded as ground, as they sit marginally below the grown ground height, and briefly mislabeled for several scan lines until the guards intervene.

Figures~\ref{fig:qual-c} and \ref{fig:qual-d} show the road method from a top view. Figure~\ref{fig:qual-c} illustrates why precision reaches 0.969: there are virtually no false positives even where only a low curb bounds the road, and both sides of the occlusion cast by a lead vehicle are seeded correctly. It also exposes the first limitation: beyond about 60\,m the scan lines are around 10\,m apart in range, and the propagation can no longer separate the road's natural slope from a raised curb, so very distant curves are dropped as false negatives. Additionally, Fig.~\ref{fig:qual-d} shows the method's principal limitation: at a crosswalk adapted with a gentle wheelchair-friendly ramp instead of a curb, the absence of a vertical boundary lets the road bleed sideways into false positives; the step-cut and cohesion pass contain most of it, but curves that cross the ramp directly cannot be distinguished from the road camber and are wrongly labeled as road together with a strip of sidewalk.

\section{Conclusion}
\label{sec:conclusion}

The automotive industry is finally integrating LiDAR into production vehicles. However, the sensors it has chosen (hybrid solid-state and solid-state) differ by design from the spinning mechanical LiDARs behind today's autonomous-driving research and public datasets. Hybrid solid-state LiDARs produce ring-free, dense, cluttered, front-facing data that challenges both the classical geometric pipelines and the learned methods built for those sensors. Yet, that same dense, focused, per-curve structure is also an opportunity, unlocking new solutions to autonomous-driving problems where the sparse data of spinning LiDAR falls short. Precise ground and road segmentation is one such problem. In hybrid solid-state data, the road boundary reveals itself through two clean geometric signatures: a height step and a range fold-back. GHySSeg exploits these signatures with a purely geometric pipeline, composed of a per-curve walk with seeded growth, multiseed propagation, and several guard mechanisms, to segment both ground and road from a single frame, with no training and no GPU, in real time.

On the benchmark that had to be assembled for this work given the unavailability of hybrid solid-state benchmarks for ground and road segmentation, GHySSeg outperforms the state of the art on both tasks, ground and road segmentation, achieving 0.896 IoU at 0.961 precision for ground and 0.907 IoU at 0.969 precision for road, ahead of both leading geometric and non-retrained learned baselines, and comfortably so on road. Precision (the safety-critical metric the pipeline is built around) is the primary result, as it keeps real obstacles from being mislabeled as traversable, a failure that can be fatal. Furthermore, the method has proven to be transferable between vehicles and sensor poses under frozen parameters without accuracy loss. Just as important, the full ground-and-road pipeline runs at $\sim$22\,ms per frame on a single CPU core, well within the sensor's 100\,ms period, demonstrating its real-time, on-board capabilities and leaving ample headroom for downstream autonomous-driving modules.

Beyond the method, this work contributes a benchmark. We release HySoLiD-Drive (100 semantically labeled point clouds and 7000 unlabeled ones from two continuous drives, recorded by two vehicles that mount the same Seyond Falcon Kinetic LiDAR) as the first public benchmark dedicated to ground and road segmentation on an automotive hybrid solid-state LiDAR, with an open leaderboard and evaluation tools. It closes a real gap: while nearly every major public autonomous-driving dataset remains mechanical-centered, the industry moves toward solid-state sensing.

The thorough evaluation of the method also surfaced two honest, bounded limitations. Where no physical boundary (a curb, guardrail, or even a drainage channel) delimits the road, the method cannot stop growing and produces false positives; and beyond roughly 60\,m, the non-uniform density of a front-facing solid-state sensor leaves far curves too sparse to classify, yielding false negatives. Future work targets both: the road-edge signatures can be complemented with the sensor's native dense and robust intensity returns to detect painted lane markings where no other physical edge exists, and temporal multi-frame accumulation via LiDAR odometry can densify distant range to recover far road, trading the single-shot property for recall. Finally, as labeled hybrid or solid-state data from other sensors becomes publicly available, it would be interesting to explore how GHySSeg transfers to their different scan patterns and point distributions, extending the cross-vehicle generalization shown here to a cross-sensor setting.

\section*{Acknowledgment}

The authors gratefully acknowledge Universidad Carlos III de Madrid for providing access to the two research autonomous vehicles used in this work and in the creation of HySoLiD-Drive.

\bibliographystyle{IEEEtran}
\bibliography{references}

\begin{IEEEbiography}[{\includegraphics[width=1in,height=1.25in,clip,trim={0cm 0cm 0cm 0cm}, keepaspectratio]{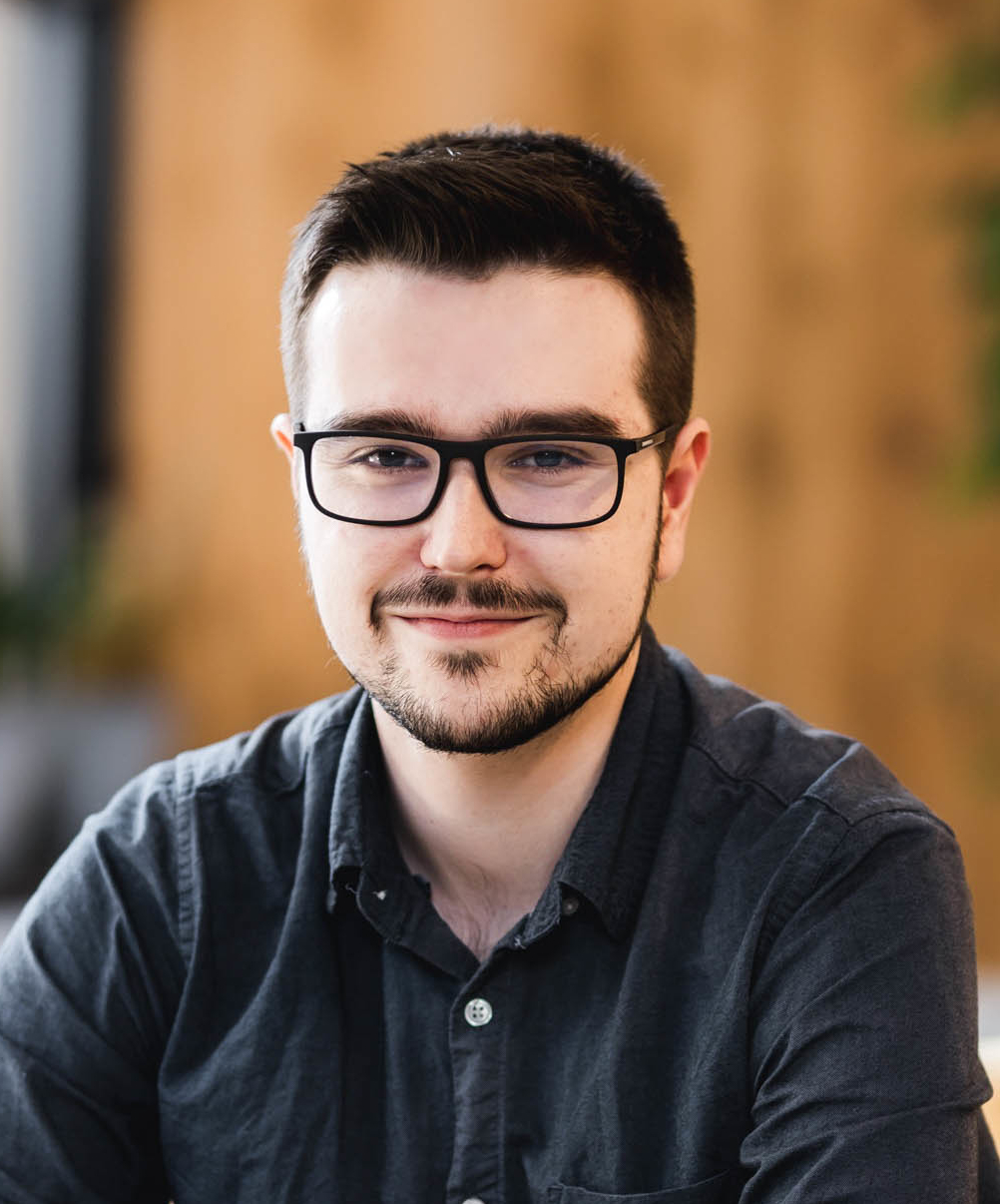}}]{Martín Palos Lorite} (Student Member, IEEE) received both his B.S. in Industrial Electronics and Automation Engineering and his M.S. in Robotics and Automation from Universidad Carlos III de Madrid (UC3M), where he is presently completing his Ph.D. in Embedded Autonomous Driving Systems. Since 2019, he has developed his career in the autonomous mobility field, contributing to projects ranging from early‑stage research and controlled real‑road demonstrations to complex industrial applications. He now serves as Technical Lead at a UC3M spin‑off company, leading the autonomous driving engineering division and overseeing the development pipeline from system design to real‑world deployment.
\end{IEEEbiography}

\begin{IEEEbiography}[{\includegraphics[width=1in,height=1.25in,clip,trim={0cm 0cm 0cm 0cm}, keepaspectratio]{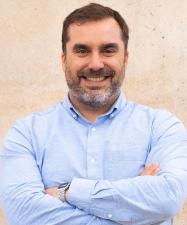}}]{Fernando García Fernández} (Member, IEEE) received the Ph.D. degree in electrical, electronic engineering and automation from the Universidad Carlos III de Madrid, Spain in 2012, where he has been serving as an Associate Professor since 2018. His research interests include perception, digital twin, and data fusion with primary focus on applications in vehicles and robotics. In 2021, he was honored with the IEEE Intelligent Transportation Systems Young Researcher/Engineer Award. He has been a member of the Board of Governors of the IEEE Intelligent Transportation Systems Society since 2017 and served as a Chair of the Spanish Chapter from 2019 to 2020.
\end{IEEEbiography}

\vfill

\end{document}